\documentclass[runningheads]{llncs}
\usepackage{graphicx}
\usepackage{comment}
\usepackage{amsmath,amssymb} 
\usepackage{color}

\begin{document}
\pagestyle{headings}
\mainmatter
\def\ECCVSubNumber{2374}  

\title{
Negative Pseudo Labeling\\ using Class Proportion\\ for Semantic Segmentation in Pathology
} 

\titlerunning{Negative Pseudo Labeling using Class Proportion for Sem. Seg. in Pathology}
%
\newcommand{\repeatthanks}{\textsuperscript{\thefootnote}}

\author{Hiroki Tokunaga\inst{1}\thanks{Contributed Equally} \and
Brian Kenji Iwana\inst{1}\repeatthanks \and
Yuki Teramoto\inst{2} \and
Akihiko Yoshizawa\inst{2} \and
Ryoma Bise\inst{1,3}}
\authorrunning{H. Tokunaga, B.K. Iwana, and R. Bise  et al.}
%
\institute{Kyushu University, Fukuoka, Japan \\
\email{\{iwana,bise\}@ait.kyushu-u.ac.jp}
\and
Kyoto University Hospital, Kyoto, Japan \\
\and
Research Center for Medical Bigdata, National Institute of Informatics, Japan
}
\maketitle

\begin{abstract}
In pathological diagnosis, since the proportion of the adenocarcinoma subtypes is related to the recurrence rate and the survival time after surgery, the proportion of cancer subtypes for pathological images has been recorded as diagnostic information in some hospitals. In this paper, we propose a subtype segmentation method that uses such proportional labels as weakly supervised labels. If the estimated class rate is higher than that of the annotated class rate, we generate negative pseudo labels, which indicate, ``input image does not belong to this negative label,'' in addition to standard pseudo labels. It can force out the low confidence samples and mitigate the problem of positive pseudo label learning which cannot label low confident unlabeled samples.
Our method outperformed the state-of-the-art semi-supervised learning (SSL) methods.
\end{abstract}

\section{Introduction}
Automated segmentation of cancer subtypes is an important task to help pathologists diagnose tumors since it is recently known that the proportion of the adenocarcinoma subtypes is related to the recurrence rate and the survival time after surgery~\cite{YoshizawaA2011}.
Therefore, the proportional information of cancer subtypes has been recorded as the diagnostic information with pathological images in some advanced hospitals. 
To obtain an accurate proportion, in general, segmentation for each subtype region should be required.
However, a pathologist does not segment the regions since it is time-consuming, and thus they roughly annotate the proportion.
The ratio of the subtypes fluctuates depending on pathologists due to subjective annotation and this proportional annotation also takes time.
Therefore, automated semantic segmentation for cancer subtypes is required.

Many segmentation methods for segmenting tumor regions have been proposed.
Although the state-of-the-art methods accurately distinguish regions in digital pathology images~\cite{hou2016patch}, most methods reported the results of binary segmentation (normal and tumor) but not for subtype segmentation. 
Although some methods~\cite{tokunagaH2019} have tackled subtype segmentation tasks, the performance is not enough yet. We consider that this comes from the lack of sufficient training data; insufficient data cannot represent the complex patterns of the subtypes.
To obtain sufficient training data, it requires the expert's annotations, and annotation for subtype segmentation takes much more time than binary segmentation.
In addition, since the visual patterns of subtypes have various appearances depending on the tissues ({\it e.g.,} lung, colon), staining methods, and imaging devices, we usually have to prepare a training data-set for each individual case.
Although it is considered that the pre-recorded proportional information will help to improve the segmentation performance, no methods have been proposed that use such information.

Let us clarify our problem setup with the proportional information that is labeled to each whole-slide images (WSI), which are widely used in digital pathology. 
To segment subtype regions, a WSI cannot be inputted to a CNN due to the huge size ({\it e.g.}, 100,000 $\times$ 50,000 pixels). Thus, most methods take a patch-based classification approach that first segments a large image into small patches and then classifies each patch~\cite{hou2016patch}.
In this case, the proportional rates of each class can be considered as the ratio between the number of patch images whose label is the same class and the total number of patches. This proportional label with a set (bag) of images (instances) can be considered as a weak-supervision.
In our problem setup, a small amount of supervised data and a large amount of weakly-supervised data are given for improving patch-level subtype classification.

Semi-supervised learning (SSL) is a similar problem setup in which a small amount of supervised data and a large amount of unlabeled data are given.
It is one of the most promising approaches to improve the performance by also using unlabeled data.
One of the common approaches of SSL is a pseudo label learning that first trains with a small amount of labeled data, and then the confident unlabeled data, which is larger than a predefined threshold, is added to the training data and the classifier is trained iteratively~\cite{lee2013pseudo}.
This pseudo labeled learning assumes that the high confidence samples gradually increase with each iteration. 
However, the visual features in test data may be different from those in the training data due to their various appearance patterns, and thus many samples are estimated with low confidence.
In this case, pseudo labels are not assigned to such low confidence samples and it does not improve the confidences ({\it i.e.}, the number of pseudo samples does not increase).
Therefore, the iteration of pseudo labeling does not improve much the classification performance.
If we set a lower confidence threshold to obtain enough amount of pseudo labels, the low confidence samples may contain the samples belong to the other class (noise samples), and it adversely affects the learning.
If we naively use the proportional labels for this pseudo learning to optimize the threshold, this problem still remains since the order of confidence in low confidence is not accurate.

In this paper, we propose a negative pseudo labeling that generates negative pseudo labels in addition to positive pseudo labels (we call the standard pseudo label as a {\it positive pseudo label} in this paper) by using the class proportional information. A negative label indicates, ``input image does not belong to this negative label.''
The method first estimates the subtype of each patch image (instance) in each WSI (bag), and then computes the estimated label proportion for each bag. 
If the estimated class rate is higher than that of the annotated class rate ({\it i.e.}, the samples of this class are over-estimated), we generate the negative pseudo labels for this class to the low confidence examples in addition to the standard pseudo labels.
These positive and negative pseudo labels are added to the training data and the classifier is trained iteratively with increasing the pseudo labels.
Our method can force out the low confidence samples in over-estimated classes on the basis of the proportional label, and it mitigates the problem of positive pseudo label learning that cannot label low confident unlabeled samples.
Our main contributions are summarized as follows:
\begin{itemize}
\vspace{-1mm}
    \item We propose a novel problem that uses proportional information of cancer subtypes as weak labels for semi-supervised learning in a multi-class segmentation task. This is a task that occurs in real applications.
    \item We propose Negative Pseudo Labeling that uses class proportional information in order to efficiently solve the weakly- and semi-supervised learning problem. Furthermore, Multi Negative Pseudo Labeling improves the robustness of the method in which it prevents the negative learning from getting hung up on obvious negative labels. 
    \item We demonstrate the effectiveness of the proposed method on a challenging real-world task. Namely, we perform segmentation of subtype regions of lung adenocarcinomas. The proposed method outperformed other state-of-the-art SSL methods.
\end{itemize}

\section{Related works}
\noindent {\bf Segmentation in pathology:} 
As data-sets for pattern recognition in pathology have opened, such as Camelyon 2016~\cite{Camelyon2016}, 2017~\cite{Camelyon2017}, many methods have been proposed for segmenting tumor regions from normal regions.
As discussed above, most methods are based on a patch-based classification approach that segments a large image into small patches and then classifies each patch separately, since a WSI image is extremely large to be inputted into a 
CNN~\cite{AltunbayD2010,ChangH2014,CruzRoaA2014,MousaviHS2015,XuY2015,ZhouY2014,WangD2016,HouL2016}. 
These methods use a fixed size patch image and thus they only use either the context information from a wide field of view or high resolution information but not both.
To address this problem, multi-scale based methods have been proposed~\cite{AlsubaieN2018,KongB2017,SirinukunwattanaK2018,tokunagaH2019,takahamaS2019}. Tokunaga {\it et al.}~\cite{tokunagaH2019} proposed an adaptively weighting multi-field-of-view CNN that can adaptively use image features from different-magnification images. Takahama {\it et al.}~\cite{takahamaS2019} proposed a two-stage segmentation method that first extracts the local features from patch images and then aggregates these features for the global-level segmentation using U-net.
However, these supervised methods require a large amount of training data in order to represent the various patterns in a class, in particular, in the multi-class segmentation.

\noindent {\bf Negative label learning:}
Negative label learning~\cite{ishida2018complementarylabel,kim2019nlnlnegative}, also called complementary label learning, is a new machine learning problem that was first proposed in 2017~\cite{ishida2017learning}.
In contrast to the standard machine learning that trains a classifier using positive labels that indicate ``input image belongs to this label,'' negative label learning trains a classifier using a negative label that indicates ``input image does not belong to this negative label.'' Ishida {\it et al.}~\cite{ishida2017learning,ishida2018complementarylabel} proposed loss functions for this negative label learning in general problem setup. Kim {\it et al.}~\cite{kim2019nlnlnegative} uses negative learning for filtering noisy data from training data. To the best of our knowledge, methods that introduce negative learning into semi- or weakly-supervised learning have not been proposed yet.

\noindent {\bf Semi-supervised learning (SSL):} 
SSL is one of the most promising approaches to improve the performance of classifiers using unlabeled data.
Most SSLs take a pseudo label learning approach~\cite{berthelot2019mixmatch,lee2013pseudo,kikkawa2019ssl,sohn2020fixmatch}. For example, pseudo labeling~\cite{lee2013pseudo} (also called self-training~\cite{xie2019self}) first trains using labeled data and then confident unlabeled data, which is larger than a predefined threshold, is added to the training data and the classifier is trained iteratively.
Consistency regularization~\cite{miyato2018virtual} generates pseudo labeled samples on the basis of the idea that a classifier should output the same class distribution for an unlabeled example even after it has been augmented.
These methods generate only positive pseudo labeled samples but not negative pseudo labels. 

\noindent {\bf Learning from label proportions (LLP):} 
LLP is the following problem setting: given the label proportion information of bags that contains a set of instances, estimating the class of each instance that is a member of a bag. 
Rueping {\it et al}~\cite{RuepingS2010} proposed a LLP that uses a large-margin regression by assuming the mean instance of each bag having a soft label corresponding to the label proportion. Kuck {\it et al}~\cite{KuckH2005} proposed a hierarchical probabilistic model that generates consistent label proportions, in which similar idea were also proposed in \cite{MusicantDR2007,ChenBC2006}. 
Yu {\it et al}~\cite{yu2013proptosvm} proposed $\propto$SVM that models the latent unknown instance labels together with the known group label proportions in a large-margin framework. SVM-based method was also proposed in \cite{QiZ2017}.
These methods assume that the features of instances are given {\it i.e.}, they cannot learn the feature representation.
Recently, Liu {\it et al}~\cite{LiuJ2019} proposed a deep learning-based method that leverages GANs to derive an effective algorithm LLP-GAN and the effectiveness was shown in using open dataset such as MNIST and CIFAR-10. The pathological image has more complex features compared with such open dataset and their method does not introduce the semi-supervised learning like fashion in order to represent such complex image features.

Unlike these current methods, we introduce negative labeling for semi-supervised learning using label proportion. 
It can effectively use the supervised data for LLP and mitigates the problem of the standard pseudo labeling that cannot label low confident unlabeled samples.

\section{Negative pseudo labeling with label proportions}

In the standard Pseudo Labeling, if the maximum prediction probability (Confidence) of an input image is higher than a set threshold, a pseudo label is assigned to the image. 
On the other hand, in the problem setting of this study, the class ratio (cancer type ratio) in the pathological image is known. 
In this section, we propose a new pseudo labeling method using this information.

\subsection{Pseudo Labeling}

Pseudo Labeling~\cite{lee2013pseudo} is an SSL technique that assigns a pseudo label to an unlabeled input pattern if the maximum prediction probability (confidence) exceeds a threshold.  
This is based on the assumption that a high probability prediction is a correct classification result and can be used to augment the training patterns. 
It functions by repeated steps of learning from supervised data, classifying unlabeled data, and learning from the new pseudo labeled data.  
Notably, the advantage of using Pseudo Labeling is that it only carries the aforementioned assumption about the data and can be used with classifiers easily. 

However, the downside of Pseudo Labeling is that it requires the unlabeled training data to be easily classified~\cite{oliver2018realistic}. 
Specifically, difficult data that does not exceed the threshold is not assigned pseudo labels and data that is misclassified can actively harm the training. 
In addition, if the distributions of the supervised data and the unsupervised data are significantly different, Pseudo Labeling cannot perform accurate pseudo labeling and will not improve discrimination performance.

\subsection{Negative Pseudo Labeling}

Pathological images differ greatly in image units even in the same cancer cell class. 
Therefore, if there is only a small number of supervised data, the distribution of supervised data and unsupervised data is significantly different and the confidence is low, so Pseudo Labeling may not be performed. Therefore, we propose Negative Pseudo Labeling, which assigns a pseudo label to data with low confidence by utilizing the cancer type ratio.

\begin{figure}[t]
        \centering
        \includegraphics[width=\linewidth]{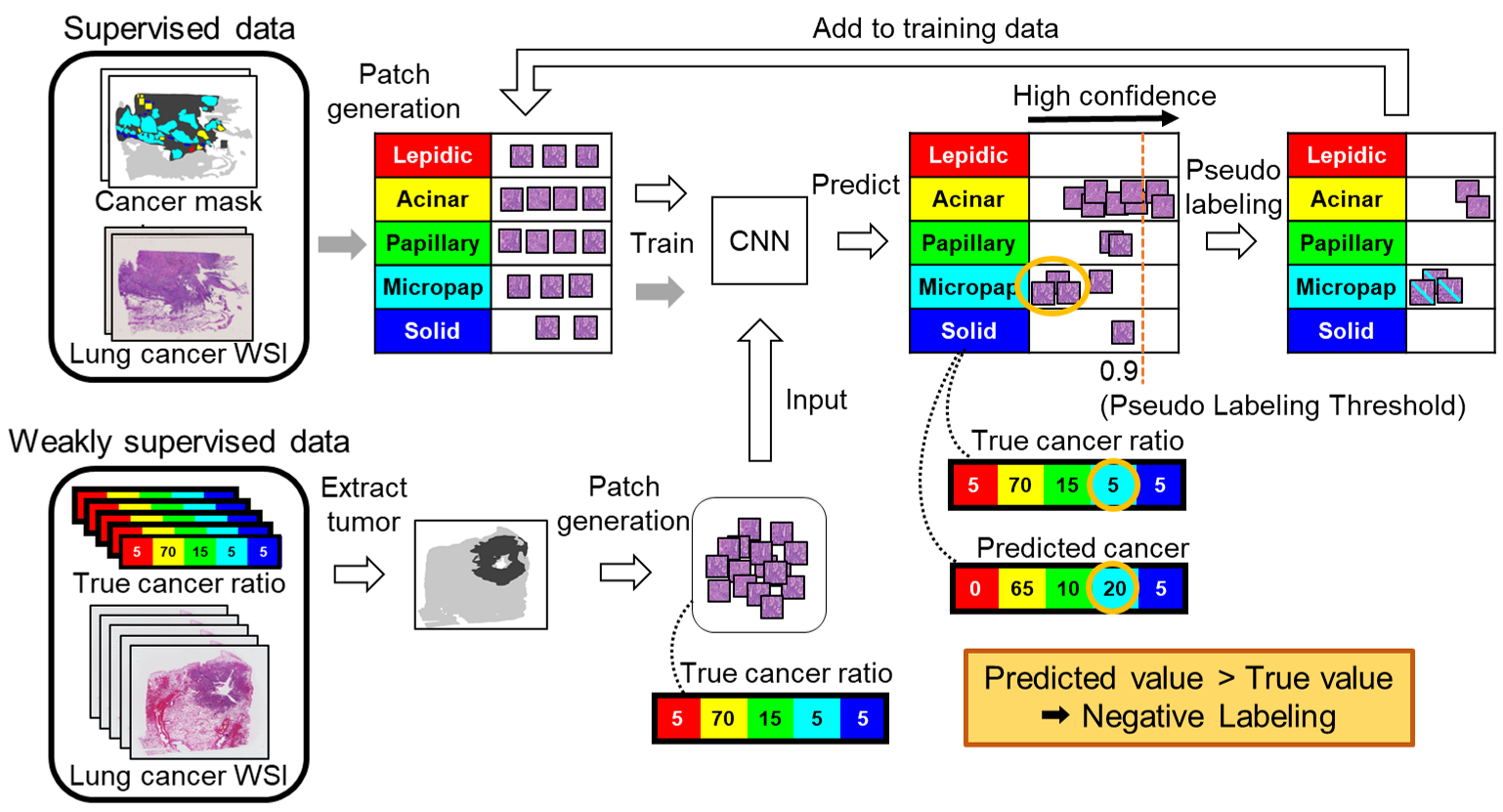}
        \caption{Overview of Negative Pseudo Labeling}
        \label{fig:negative_pseudo_labeling}
    \end{figure}

In Negative Pseudo Labeling, in addition to normal pseudo labeling, negative pseudo labels are added to incorrectly predicted images using weak teacher information of cancer type ratio.
Figure~\ref{fig:negative_pseudo_labeling} shows the outline of Negative Pseudo Labeling.
First, a training patch image is created from a small number of supervised data, and the CNN is trained (gray arrow, executed only the first time).
Next, a tumor region is extracted from weakly supervised data, and a patch image is created from that region.
At this time, a set of patch images created from one pathological image has weak teacher information of the cancer type ratio.
Next, the patch image is input to the CNN trained with supervised data, and the class probability is predicted.
In the case of normal pseudo labeling, pseudo labels are given only to patches whose confidence is greater than or equal to the threshold. 
In the proposed method, negative pseudo labels are assigned to those with less than the threshold using cancer type ratio information.
    
By applying pseudo labeling not only for patch images but for whole pathological images, the predicted value of the cancer type ratio can be calculated.
The correct answer and the predicted cancer type ratio are compared for each class. 
If the predicted value is larger than the correct answer value, a negative label is assigned to the corresponding image because the class is excessively incorrectly predicted.
For example, in Figure~\ref{fig:negative_pseudo_labeling}, the cyan class (Micro papillary) was predicted to be 20\%.
However, the correct ratio is 5\%. 
Thus, the excess 15\% in reverse order of per patch confidence is added as a negative label. 
This pseudo teacher is added to the training data, and the model is trained again.
These steps are repeated until the loss of the validation data converges.
    
Specifically, the assignment of the actual negative pseudo label is performed according to the Negative Selectivity (NS) calculated by:
\begin{equation} \label{eq:negative_selectivity}
    NS_c = SAE \times max ( 0,~PCR_c - TCR_c) \times ( 1 - TCR_c/PCR_c),
\end{equation}
where $c$ is the class, the sum absolute error $SAE$ is the distance between the ratios, and $TCR_c$ and $PCR_c$ are the true cancer ratio and predicted cancer ratio, respectively. 
SAE is defined by the following equation and indicates the sum of errors between the ratios of one pathological image in the range of 0 to 1:
\begin{equation} \label{eq:sae}
    SAE = max(1, \sum_{c=1}^{C} |TCR_c - PCR_c|).
\end{equation}
Here, $C$ represents the number of classification classes.
The first term of Eq.~\eqref{eq:negative_selectivity} selects pathological images with many incorrect predictions, the second term gives negative pseudo labels to only excessively incorrect classes, and the third term gives negative pseudo labels to classes that are close to 0\%.

\begin{figure}[t]
    \centering
    \includegraphics[width=\linewidth]{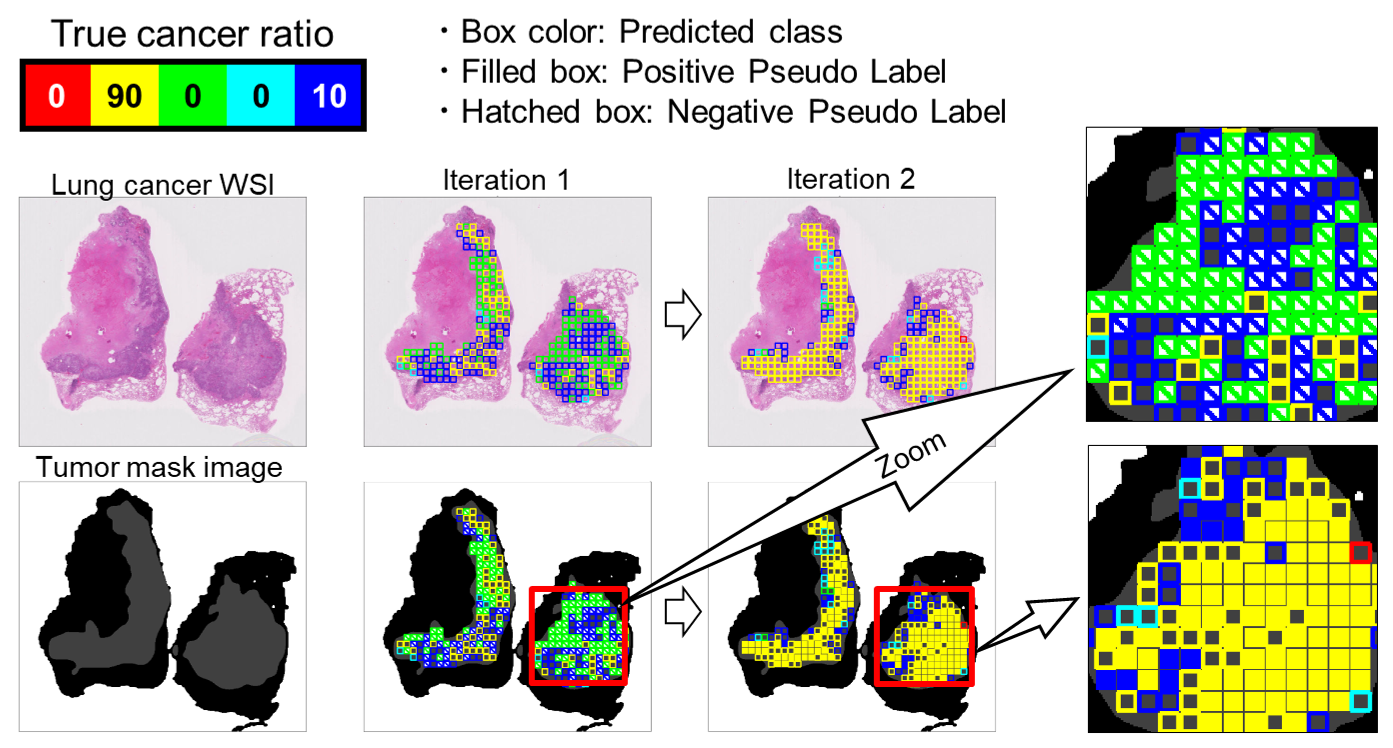}
    \caption{Example of Negative Pseudo Label assignment. The color of the box is the predicted class, the filled boxes are patches with positive pseudo labels, and the hatched boxes are the patches with negative pseudo labels. By learning from the negative pseudo label, the prediction class changes to a different class from green.}
    \label{fig:labeling_sample}
\end{figure}

In the proposed method, in addition to the positive pseudo label assigned by normal Pseudo Labeling, the new negative pseudo labels are added to enable effective learning.
Fig.~\ref{fig:labeling_sample} shows an example of a negative and positive pseudo label.
From the ratio of correct cancer types, it can be seen that there should be none classified as green and 10\% blue. 
However, when observing the predicted class in Iteration~1, it can be seen that blue and green occupy the majority classifications. 
Thus, they are incorrectly over-predicted. 
Therefore, a negative pseudo label (hatched box) is assigned to the patches in NS order. 
In Iteration~2, learning from the pseudo labels is performed and new predictions changed to the yellow class. 
In other words, negative pseudo labels are assigned to a patch that has been predicted as a class that should not exist which causes the model to change toward the correct class. 

\subsection{Multi Negative Pseudo Labeling}
\begin{figure}[t]
\centering
\includegraphics[width=0.9\linewidth]{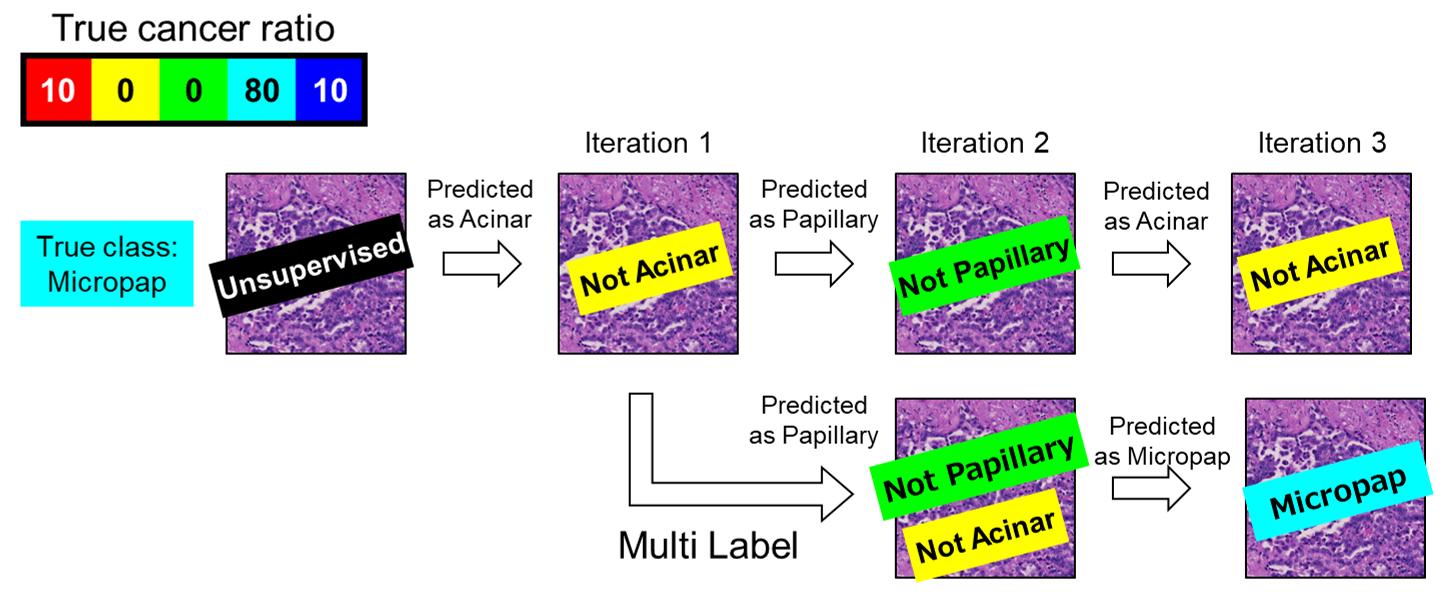}
\caption{Example of a negative label oscillating between wrong classes. By storing past label information, this can be prevented.}
\label{fig:nl_fluctuation}
\end{figure}

By giving a negative pseudo label, it is possible to change the prediction from one class to another. 
However, in negative learning, the new class is not necessarily the correct class. 
Fig.~\ref{fig:nl_fluctuation} shows an example in which the learning does not progress because the negative label oscillates between the wrong classes.
In the figure, the true correct class is Micro papillary, but only the weakly supervised ratio is known.
When Negative Pseudo Labeling was performed, the patch was predicted to be Acinar, which was judged to be an incorrect prediction class based on the cancer type ratio.
Therefore, a negative label of Acinar is assigned in Iteration~1.
When the class was predicted again by Iteration~2, this time, the prediction was changed to another class which should not exist, Papillary. 
By repeating Negative Pseudo Labeling, this situation can oscillate between incorrect classes. 
In the case of Negative Pseudo Labeling, the loss function,
\begin{equation} \label{n_loss}
    \mathcal{L}_{negative} = -\sum_{c=1}^{C} \boldsymbol{y}_{c} \log \left(1-\log {f(c|\boldsymbol{x},\theta)}\right),
\end{equation}
can get stuck by only minimizing alternating predictions between erroneous classes. 
    
Therefore, Negative Pseudo Labeling is extended to Multi Negative Pseudo Labeling to prevent these oscillations.
In Negative Pseudo Labeling, positive and negative pseudo labels are added for each Iteration, and the past pseudo label information is not used.
Therefore, we accumulate the pseudo label information given by each iteration and give a multi-negative pseudo label.
However, simply adding negative labels increases the scale of the loss, therefore, we propose a loss function $\mathcal{L}_{multi-negative}$ that performs weighting based on the number of negative labels, or:
\begin{equation} \label{n_loss_multi}
    \mathcal{L}_{multi-negative} = - \frac{(C - |\boldsymbol{{y}}|)}{(C-1)}  \sum_{c=1}^{C} \boldsymbol{y}_{c} \log \left(1-\log {f(c|\boldsymbol{x},\theta)}\right),
\end{equation}
where $\boldsymbol{{y}}_c$ is a negative label for a sample, $|\boldsymbol{{y}}|$ is the number of negative labels for a sample on the iteration, $\boldsymbol{x}$ is the input data, $f(c|\boldsymbol{x}, \theta) $ is the predicted probability of class $c$, and $C$ is the number of classification classes.
The loss $\mathcal{L}_{multi-negative}$ is the same with the loss (Eq. \ref{n_loss}) for the single negative label when $|\boldsymbol{y}|$ is $1$ ({\it i.e.,} a single negative pseudo label is given in the iteration), and the weight linearly decreases if the number of negative pseudo labels ($|\boldsymbol{y}|$) increases.

\section{Adaptive Pseudo Labeling}

Conventionally, Pseudo Labeling ignores the unlabeled data with low prediction probabilities and through the introduction of Negative Pseudo Labeling, it now is possible to use these inputs. 
However, in order to assign positive pseudo labels, it still is necessary to manually set a threshold value above which the positive pseudo label is assigned. 
In this section, we propose the use of Adaptive Pseudo Labeling, which adaptively determines the criteria for positive pseudo labeling based on the similarity of cancer types.
    
\begin{figure}[t]
        \centering
        \includegraphics[width=0.85\linewidth]{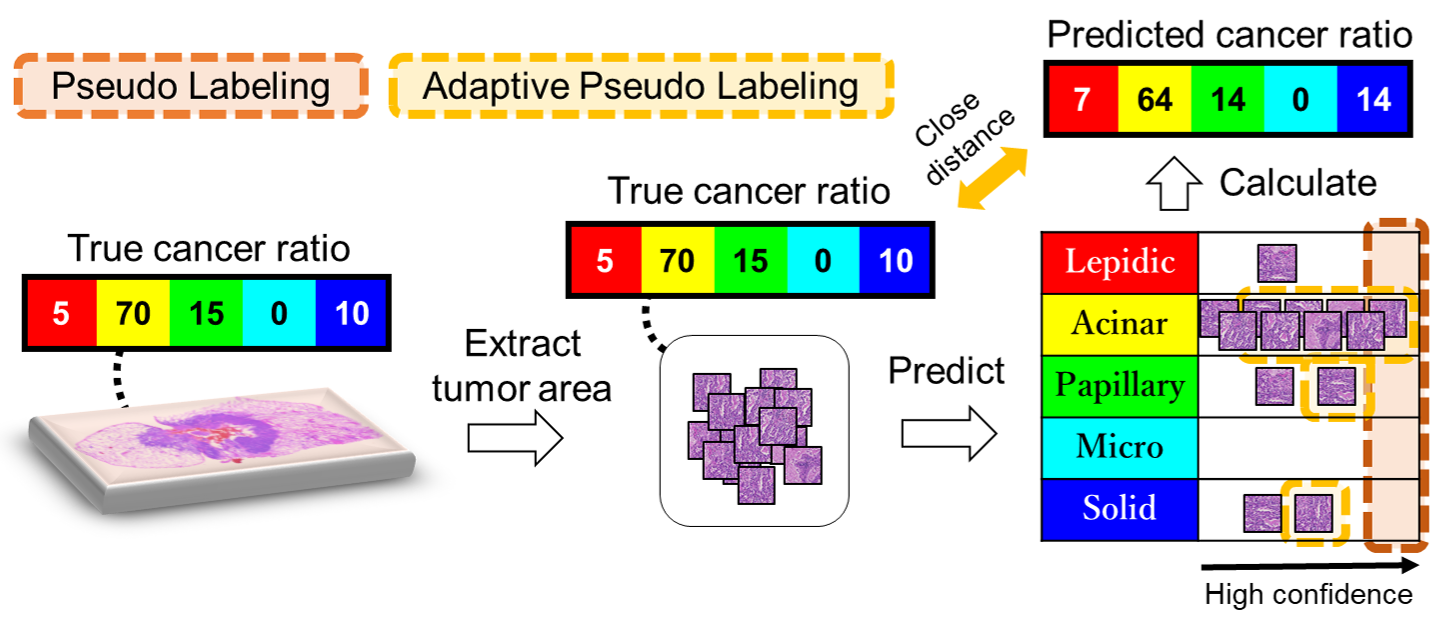}
        \caption{Example of cancer type ratio information. When accurate prediction is performed, the similarity between the predicted cancer type ratio of the set of patch images and the correct cancer type ratio increases.}
        \label{fig:cancer_ratio_sample}
    \end{figure}

When an accurate prediction is made for a patch image and the predicted cancer type ratio is close to the correct cancer type ratio, we consider the similarity between the ratios as large.
For example, in Figure~\ref{fig:cancer_ratio_sample}, the correct cancer type ratio is [5\%, 70\%, 15\%, 0\%, 10\%] and the predicted cancer type ratio is [7\%, 64\%, 14\%, 0\%, 14\%], the similarity between the ratios is high.
However, in a single pathological image, it is not possible to judge that the class label of each patch image is accurate just because the similarity between the ratios is high.
In Figure~\ref{fig:cancer_ratio_sample}, even if the green and blue predicted patch images are switched, the predicted cancer type ratio remains at $14\%$ and the similarity does not change.
However, if there are multiple pathological images and the ratio of cancer types differs for each pathological image, it is unlikely that the ratios will be the same when predictions between different classes are switched.
Therefore, the prediction class of the patch image is considered reliable when the similarity of the cancer type ratio is high.
    
This is represented as the Positive Selectivity~(PS) of the patch being assigned a pseudo label. 
PS is a parameter that determines how much of the images are given a positive pseudo label, and is defined by:
    \begin{equation} \label{eq:positive_selectivity}
        PS_c = (1 - SAE) \times min( TCR_c, PCR_c),
    \end{equation}
where $SAE$ is the distance between the ratios defined by Eq.~\eqref{eq:sae}, and $TCR_c$ and $PCR_c$ are the true cancer ratio and predicted cancer ratio, respectively.
Eq.~\eqref{eq:positive_selectivity} is calculated for each class of pathological image, and positive pseudo labels are assigned to only $PS_c$ total patch images in the order of confidence.
    
Figure~\ref{fig:cancer_ratio_sample} shows an example of Adaptive Pseudo Labeling.
In Pseudo Labeling, pseudo labeling is performed with a fixed threshold regardless of the distance of the cancer type ratio, so it is typical of pseudo labeling being only applied to a small number of images.
On the other hand, in Adaptive Pseudo Labeling, the selection rate is dynamically determined for each pathological image according to the distance of the ratio. 


\section{Experimental Results}
\label{result}


\subsection{Dataset} 
    
\begin{figure}[t]
    \centering
    \includegraphics[width=\linewidth,clip,trim={2.7cm 9.7cm 0.3cm 4.7cm}]{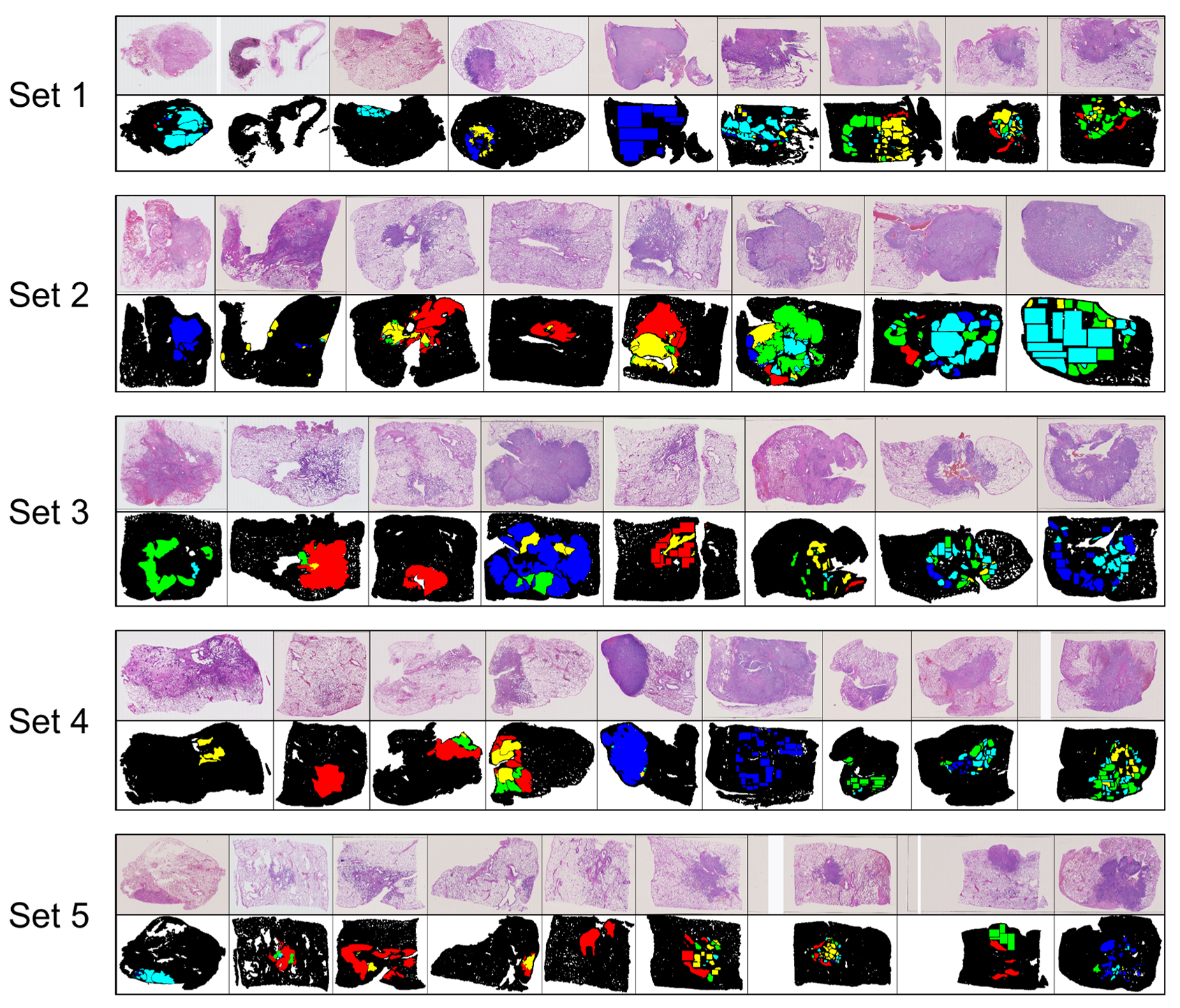}
    \caption{Examples of 16 of the 42 pathological images and the ground truth regions. Red, yellow, green, cyan, and blue indicate Lepidic, Acinar, Papillary, Micro papillary, and Solid, respectively.}
    \label{fig:5foldmask}
\end{figure}

In the experiment, we use a real-world dataset consisting of 42 supervised pathological images and 400 weakly supervised pathological images. 
Each pathological image contains a slide of a tumor that is up to $108,000\times54,000\times3$ pixels in size and was annotated by two pathologists.
The supervised images have pixel-wise class labels and the weakly supervised images only have a general ratio of cancer types. 
As shown in Fig.~\ref{fig:5foldmask}, the supervised images are annotated into one non-cancer region and five cancer regions, Lepidic, Acinar, Papillary, Micro papillary, and Solid. 
In addition, there are vague areas in the tumor area, indicated in black, which were difficult for the pathologists to classify. 
These difficult regions were excluded when evaluating the accuracy of the cancer type segmentation. 
As for the weakly supervised images, the pathologists only provide an estimated ratio of cancer type in percent.

In order to construct patch images, the WSI images are broken up using a non-overlapping sliding window of size $224\times224\times3$ pixels at a magnification of $10\times$ (a typical magnification used for diagnosis). 
As a result, there are a total of 5,757, 2,399, 3,850, 3,343, and 4,541 labeled patches for Lepidic, Acinar, Papillary, Micro papillary, and Solid cancer cells, respectively. 
These patches are separated five pathological image-independent sets for 5-fold cross-validation. 
Specifically, each cross-validation labeled training set is created from patches of 33 to 34 of the original full-size images with 8 to 9 of the full-size images left out for the test sets. 
From each labeled training set, 25\% random the patch images are taken and used as respective validation sets. 

To extract patches for the pseudo labeled training set, patches of tumor regions were extracted from the weakly supervised data. 
Due to the weakly supervised data only containing the ratio of cancer types and not the full annotations, we used a Fix Weighting Multi-Field-of-View CNN~\cite{tokunagaH2019} to extract tumor proposal regions in order to create patches. 
The Multi-Field-of-View CNN combines multiple magnifications in order to accomplish segmentation. 
In this case, we used the magnifications of $10\times$, $5\times$, and $2.5\times$.
In total, 52,714 patches from the weakly supervised data are used.

\subsection{Experiment settings}
\label{experimental_setting}

For the base CNN architecture, the proposed method uses a MobileNetV2~\cite{sandler2018mobilenetv2} with the weights pre-trained by ImageNet. 
During model training, the network was fine-tuned with the convolutional layers frozen~\cite{agrawal2014analyzing}. 
The training was performed in two steps. 
First, the MobileNetV2 was trained using the supervised data for 100 epochs with early stopping based on a validation loss plateau of 10 epochs.  
Second, the network was fine-tuned using the Pseudo Labeling for a similar 100 epochs with early stopping.  
During the Pseudo Labeling step, the positive learning loss $\mathcal{L}_{positive}$,
\begin{equation} \label{p_loss}
    \mathcal{L}_{positive} = -\sum_{c=1}^{C} \boldsymbol{y}_{c} \log {f(c|\boldsymbol{x},\theta)},
\end{equation}
where $\boldsymbol{y}_{c}$ is the correct label, $\boldsymbol{x}$ is input data, and $f(\cdot)$ is a classifier with parameter $\theta$.
For the Negative Pseudo Labeling, the multi-negative learning loss $\mathcal{L}_{multi-negative}$ from \eqref{n_loss_multi} is used. 
Furthermore, the network was trained using mini-batches of size 18 with an RAdam~\cite{liu2019variance} optimizer with an initial learning rate of $10^{-5}$.

\subsection{Quantitative Evaluation}
The prediction results were quantitatively evaluated using three evaluations recommended for region-based segmentation~\cite{CsurkaG2013}, Overall Pixel~(OP), Per-Class~(PC), and the mean Intersection over Union~(mIoU). 
These metrics are defined as: 
\begin{equation}
\operatorname{OP} = \frac{\sum_{c}{TP_c}}{\sum_c{(TP_c+FP_c)}},\hspace{3mm} \operatorname{PC} = \frac{1}{M}\sum_{c}{\frac{TP_c}{TP_c+FP_c}},\nonumber
\end{equation}
\vspace{-1.5mm}
\begin{equation}
\operatorname{mIoU} = \frac{1}{M}\sum_{c}{\frac{TP_c}{TP_c+FP_c+FN_c}},
\end{equation}
where $M$ is the number of classes, and $TP_c$, $FP_c$, and $FN_c$ are the numbers of true positives, false positives, and false negatives for class $c$, respectively. 

The results of the quantitative evaluations are shown in Table~\ref{Table:comp}. 
In this table, we compare the results of using the proposed Multi Negative Pseudo Labeling with Adaptive Positive Pseudo Labeling~(Proposed) with state-of-the-art SSL methods. 
As baselines, Supervised uses only the supervised training data and Pseudo Labeling~\cite{lee2013pseudo} is the standard implementation of using positive pseudo labels with a threshold of 0.95. 
Interpolation Consistency Training (ICT)~\cite{Verma_2019}, Mean Teachers~\cite{tarvainen2017mean}, mixup~\cite{zhang2018mixup}, Virtual Adversarial Training (VAT)~\cite{miyato2018virtual} are other recent and popular SSL methods that we evaluated on our dataset. 
These methods were trained under similar conditions as Proposed. 
It should be noted that the SSL methods can only use the weakly supervised data as unlabeled data.
From the quantitative evaluation results, it can be seen that the Proposed accuracy is better than the other SSL methods. 

\begin{table}[t] 
\begin{center}
\caption{Quantitative Evaluation Results}
\label{Table:comp} 
\begin{tabular}{lccc}
\hline
Method & OP & PC & mIoU \\ 
\hline
Proposed & \textbf{0.636} & \textbf{0.588} & \textbf{0.435} \\
Supervised & 0.576 & 0.527 & 0.373 \\
Pseudo Labeling~\cite{lee2013pseudo} & 0.618 & 0.575 & 0.420 \\
ICT~\cite{Verma_2019} & 0.461 & 0.430 & 0.289 \\
Mean Teachers~\cite{tarvainen2017mean} & 0.449 & 0.421 & 0.270 \\
mixup~\cite{zhang2018mixup} & 0.529 & 0.476 & 0.324 \\
VAT~\cite{miyato2018virtual} & 0.477 & 0.427 & 0.281 \\
\hline
\end{tabular}
\end{center}
\end{table}

Furthermore, we performed ablation experiments to demonstrate the importance of the Negative Pseudo Labeling. 
These results are shown in Table~\ref{Table:result}.
In the result, the methods using only (Single/Multi) Negative Pseudo Labeling (NPL) did not work well due to class imbalance because NPL can not directly use an oversampling technique. 
When we use the Positive Labeling (PL) with NPL, the class imbalance problem can be mitigated by oversampling in PL, and thus the combination of PL and NPL improves the performance. In addition, Multi NPL further improved the performance compared with using NPL.

\begin{table}[t]
\begin{center}
\caption{Ablation Results}
\label{Table:result}
\begin{tabular}{llccc}
\hline
Positive Threshold & Negative Labeling & OP & PC & mIoU \\ 
\hline
Fixed at 0.95 & -- & 0.618 & 0.575 & 0.420 \\
Adaptive & -- & 0.613 & 0.565 & 0.412 \\
-- & Single label & 0.563 & 0.508 & 0.348 \\
-- & Multi label & 0.551 & 0.478 & 0.336 \\
Adaptive & Single label & 0.621 & 0.585 & 0.426 \\
Adaptive & Multi label & \textbf{0.636} & \textbf{0.588} & \textbf{0.435} \\ 
\hline
\end{tabular}
\end{center}
\end{table}

\subsection{Qualitative Evaluation}

\begin{figure}[t]
    \centering
    \includegraphics[width=\linewidth]{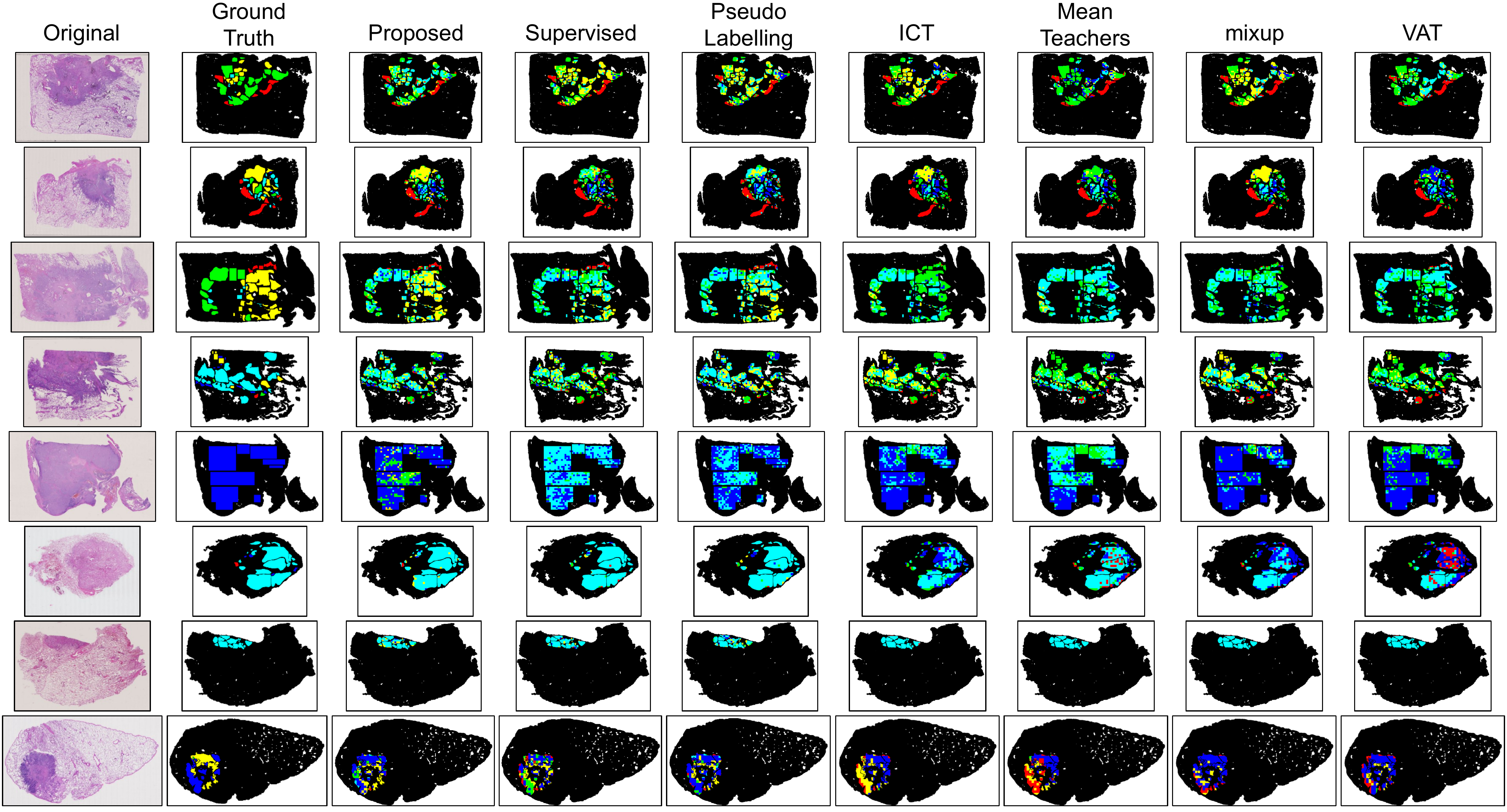}
    \caption{Results from a sample test set.}
    \label{fig:prediction}
\end{figure}

In order to evaluate the results using qualitative analysis, the patches are recombined into the full pathological images. 
Fig.~\ref{fig:prediction} shows the results from the first cross-validation test set. 
The first two columns are the original image and ground truth and the subsequent eight columns are the synthesized images with the colors corresponding to the classes listed in Fig.~\ref{fig:5foldmask}. 
The figure is able to provide an understanding of how the proposed method was able to have higher quantitative results. 
For example, when comparing the fifth row of Fig.~\ref{fig:prediction}, the SSL methods, including the proposed method, generally had much better results than just using the labeled data in Supervised. 
Furthermore, the performance improvement from using Multi Negative Pseudo Labels can clearly be seen when comparing the results from Proposed and Pseudo Labeling. 
The figure also shows that many comparison methods were able to excel at particular images, even sometimes better than the proposed method. 
However, the proposed method had more consistently accurate results for all of the images.

\begin{figure}[t]
    \centering
    \includegraphics[width=0.65\linewidth]{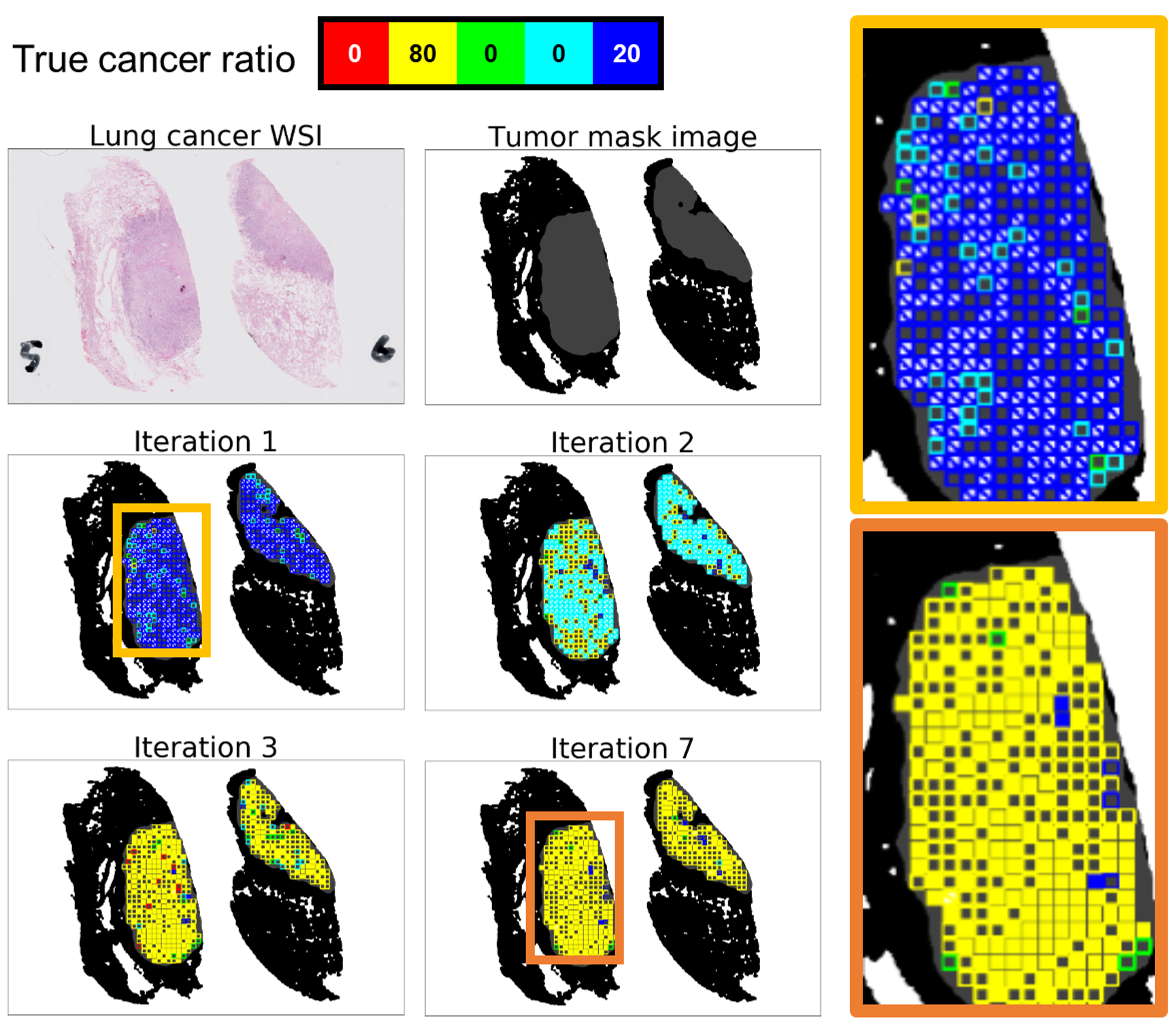}
    \vspace{-2mm}
    \caption{Example of the proposed negative and positive pseudo labeling. The color of the box is the predicted class, the filled boxes are patches with positive pseudo labels, and the hatched boxes are the patches with negative pseudo labels.}
    \label{fig:selected_patch_good_nega}
    \vspace{-2mm}
\end{figure}

In a specific instance, Fig.~\ref{fig:selected_patch_good_nega} shows the process of using positive and negative pseudo labels to direct the classifier to use the unsupervised data correctly. 
Before Iteration~1, the network is trained using only the supervised training set. 
According to the training, most of the weakly supervised patches were incorrectly classified as blue. 
Over the course of a few iterations, the pseudo labels of the patches shifted the distribution of classes to be more similar to the weakly supervised ratio. 
Before, the pseudo labels would have harmed the supervised training set. 
Now, instead, the pseudo labels are able to efficiently augment the supervised data.

\section{Conclusion}

This paper tackles cancer type classification using massive pathological images. 
A feature of using massive pathological images is that precise pixel-wise annotations are difficult and costly to acquire. 
Therefore, we proposed a new method of semi-supervised learning that incorporates positive and negative pseudo label learning. 
In this model, we introduce Negative Pseudo Labeling and its extension Multi Negative Pseudo Labeling. 
The Negative Pseudo Labeling assigns negative pseudo labels to unlabeled weakly supervised data in order to guide the pseudo label augmentation toward a ratio of classes provided by the weak supervision. 
Multi Negative Pseudo Labeling extends this idea and weights past negative pseudo labels in order to avoid getting stuck in alternating negative pseudo labels. 
Furthermore, we introduce Adaptive Pseudo Labeling for the positive pseudo labeling step, which dynamically selects the positive pseudo labels without the need of determining the threshold traditionally used in Pseudo Labeling. 
In order to evaluate the proposed method, we performed quantitative and qualitative analysis using pathological image-independent 5-fold cross-validation. 
We are able to demonstrate that the proposed method outperforms other state-of-the-art SSL methods. 
Through these promising results, we are able to show that segmentation via patch classification augmented with weak supervision is possible on real and large-scale pathological images. 

\section*{ACKNOWLEDGMENT}
This work was supported by JSPS KAKENHI Grant Number 20H04211.

\clearpage
%
%
\bibliographystyle{splncs04}
\bibliography{main}

\begin{thebibliography}{10}
\providecommand{\url}[1]{\texttt{#1}}
\providecommand{\urlprefix}{URL }
\providecommand{\doi}[1]{https://doi.org/#1}

\bibitem{agrawal2014analyzing}
Agrawal, P., Girshick, R., Malik, J.: Analyzing the performance of multilayer
  neural networks for object recognition. In: ECCV. pp. 329--344. Springer
  (2014)

\bibitem{AlsubaieN2018}
Alsubaie, N., Shaban, M., Snead, D., Khurram, A., Rajpootation, N.: A
  multi-resolution deep learning framework for lung adenocarcinoma growth
  pattern classific. In: MIUA. pp. 3--11 (2018)

\bibitem{AltunbayD2010}
Altunbay, D., Cigir, C., Sokmensuer, C., GunduzDemi, C.: Color graphs for
  automated cancer diagnosis and grading. IEEE Trans. Biomedical Engin.
  \textbf{57}(3),  665--674 (2010)

\bibitem{Camelyon2017}
Bandi, P., Geessink, O., Manson, Q., van Dijk{\it et.al.}, M.: From detection
  of individual metastases to classification of lymph node status at the
  patient level: the camelyon17 challenge. IEEE Trans. Medical Imag.  (2018)

\bibitem{Camelyon2016}
Bejnordi, B.E., Veta, M., van Diest, P.J., van Ginneken~{\it et.al.}, B.:
  Diagnostic assessment of deep learning algorithms for detection of lymph node
  metastases in women with breast cancer. Jama  \textbf{318}(22),  2199--2210
  (2017)

\bibitem{berthelot2019mixmatch}
Berthelot, D., Carlini, N., Goodfellow, I., Papernot, N., Oliver, A., Raffel,
  C.A.: Mixmatch: A holistic approach to semi-supervised learning. In: NeurIPS.
  pp. 5050--5060 (2019)

\bibitem{ChangH2014}
Chang, H., Zhou, Y., Borowsky, A., Barner, K., Spellman, P., Parvin, B.:
  Stacked predictive sparse decomposition for classification of histology
  sections. Int. J. Comput. Vis.  \textbf{113}(1),  3--18 (2014)

\bibitem{ChenBC2006}
Chen, B., Chen, L., Ramakrishnan, R., Musicant, D.: Learning from aggregate
  views. In: ICDE. pp.~3--3 (2006)

\bibitem{CruzRoaA2014}
Cruz-Roa, A., Basavanhally, A., Gonzalez, F., {\it et.al.}, H.G.: Automatic
  detection of invasive ductal carcinoma in whole slide images with
  convolutional neural networks. In: SPIE Medical Imag. (2014)

\bibitem{CsurkaG2013}
Csurka, G., Larlus, D., Perronnin, F.: What is a good evaluation measure for
  semantic segmentation? In: CVPR (2013)

\bibitem{KuckH2005}
Hendrik, K., Nando~de, F.: Svm classifier estimation from group probabilities.
  In: CUAI. pp. 332--339 (2005)

\bibitem{hou2016patch}
Hou, L., Samaras, D., Kurc, T.M., Gao, Y., Davis, J.E., Saltz, J.H.:
  Patch-based convolutional neural network for whole slide tissue image
  classification. In: CVPR. pp. 2424--2433 (2016)

\bibitem{HouL2016}
Hou, L., Samaras, D., Kurc, T.M., Gao, Y.{\it et.al.}.: Patch-based
  convolutional neural network for whole slide tissue image classification. In:
  CVPR. pp. 2424--2433 (2016)

\bibitem{ishida2018complementarylabel}
Ishida, T., Niu, G., Menon, A.K., Sugiyama, M.: Complementary-label learning
  for arbitrary losses and models. arXiv preprint arXiv:1810.04327  (2018)

\bibitem{LiuJ2019}
Jiabin, L., Bo, W., Zhiquan, Q., YingJie, T., Yong, S.: Learning from label
  proportions with generative adversarial networks. In: NeurIPS (2019)

\bibitem{kikkawa2019ssl}
Kikkawa, R., Sekiguchi, H., Tsuge, I., Saito, S., Bise, R.: Semi-supervised
  learning with structured knowledge for body hair detection in photoacoustic
  image. In: ISBI (2019)

\bibitem{kim2019nlnlnegative}
Kim, Y., Yim, J., Yun, J., Kim, J.: Nlnl: Negative learning for noisy labels.
  In: ICCV. pp. 101--110 (2019)

\bibitem{KongB2017}
Kong, B., Wang, X., Li, Z., Song, Q., Zhang, S.: Cancer metastasis detection
  via spatially structured deep network. In: IPMI. pp. 236--248 (2017)

\bibitem{lee2013pseudo}
Lee, D.H.: Pseudo-label: The simple and efficient semi-supervised learning
  method for deep neural networks. In: ICML Workshops. vol.~3, p.~2 (2013)

\bibitem{liu2019variance}
Liu, L., Jiang, H., He, P., Chen, W., Liu, X., Gao, J., Han, J.: On the
  variance of the adaptive learning rate and beyond. arXiv preprint
  arXiv:1908.03265  (2019)

\bibitem{miyato2018virtual}
Miyato, T., Maeda, S.i., Koyama, M., Ishii, S.: Virtual adversarial training: a
  regularization method for supervised and semi-supervised learning. IEEE Trans
  Pattern Analy. Mach. Intelli.  \textbf{41}(8),  1979--1993 (2018)

\bibitem{MousaviHS2015}
Mousavi, H., Monga, V., Rao, G., Rao, A.U.: Automated discrimination of lower
  and higher grade gliomas based on histopathological image analysis. J.
  Pathology Informatics  (2015)

\bibitem{MusicantDR2007}
Musicant, D., Christensen, J., Olson, J.: Supervised learning by training on
  aggregate outputs. In: ICDM. pp. 252--261 (2007)

\bibitem{oliver2018realistic}
Oliver, A., Odena, A., Raffel, C.A., Cubuk, E.D., Goodfellow, I.: Realistic
  evaluation of deep semi-supervised learning algorithms. In: NeurIPS. pp.
  3235--3246 (2018)

\bibitem{QiZ2017}
Qi, Z., Wang, B., Meng, F.: Learning with label proportions via npsvm (10),
  3293--–3305 (2017)

\bibitem{RuepingS2010}
Rueping, S.: Svm classifier estimation from group probabilities. In: ICML
  (2010)

\bibitem{sandler2018mobilenetv2}
Sandler, M., Howard, A., Zhu, M., Zhmoginov, A., Chen, L.C.: Mobilenetv2:
  Inverted residuals and linear bottlenecks. In: CVPR. pp. 4510--4520 (2018)

\bibitem{takahamaS2019}
Shusuke, T., Yusuke, K., Yusuke, M., Hiroyuki, A., Masashi, F., Akihiko, Y.,
  Masanobu, K., Tatsuya, H.: Multi-stage pathological image classification
  using semantic segmentation. In: ICCV (2019)

\bibitem{SirinukunwattanaK2018}
Sirinukunwattana, K., Alham, N., Verrill, C., Rittscher, J.: Improving whole
  slide segmentation through visual context - a systematic study. In: MICCAI.
  pp. 192--200 (2018)

\bibitem{sohn2020fixmatch}
Sohn, K., Berthelot, D., Li, C.L., Zhang, Z., Carlini, N., Cubuk, E.D.,
  Kurakin, A., Zhang, H., Raffel, C.: Fixmatch: Simplifying semi-supervised
  learning with consistency and confidence. arXiv preprint arXiv:2001.07685
  (2020)

\bibitem{ishida2017learning}
Takashi, I., Gang, N., Weihua, H., Masashi, S.: Learning from complementary
  labels. In: NeurIPS (2017)

\bibitem{tarvainen2017mean}
Tarvainen, A., Valpola, H.: Mean teachers are better role models:
  Weight-averaged consistency targets improve semi-supervised deep learning
  results. In: NeurIPS. pp. 1195--1204 (2017)

\bibitem{tokunagaH2019}
Tokunaga, H., Teramoto, Y., Yoshizawa, A., Bise, R.: Adaptive weighting
  multi-field-of-view cnn for semantic segmentation in pathology. In: CVPR
  (2019)

\bibitem{Verma_2019}
Verma, V., Lamb, A., Kannala, J., Bengio, Y., Lopez-Paz, D.: Interpolation
  consistency training for semi-supervised learning. In: IJCAI. pp. 3635--3641
  (2019)

\bibitem{WangD2016}
Wang, D., Khosla, A., Gargeya, R., Irshad, H., Beck, A.H.: Deep learning for
  identifying metastatic breast cancer. arXiv preprint arXiv:1606.05718  (2016)

\bibitem{xie2019self}
Xie, Q., Hovy, E., Luong, M.T., Le, Q.V.: Self-training with noisy student
  improves imagenet classification. arXiv preprint arXiv:1911.04252  (2019)

\bibitem{XuY2015}
Xu, Y., Jia, Z., Ai, Y., Zhang, F., Lai, M., Chang, E.I.C.: Deep convolutional
  activation features for large scale brain tumor histopathology image
  classification and segmentation. In: ICASSP (2015)

\bibitem{YoshizawaA2011}
Yoshizawa, A., Motoi, N., Riely, G.J., Sima, C.{\it et.al.}.: Impact of
  proposed iaslc/ats/ers classification of lung adenocarcinoma: prognostic
  subgroups and implications for further revision of staging based on analysis
  of 514 stage i cases. Modern Pathology  \textbf{24}(5), ~653 (2011)

\bibitem{yu2013proptosvm}
Yu, F., Liu, D., Kumar, S., Tony, J., Chang, S.F.: $\propto$svm for learning
  with label proportions. In: ICML. pp. 504--512 (2013)

\bibitem{zhang2018mixup}
Zhang, H., Cisse, M., Dauphin, Y.N., Lopez-Paz, D.: mixup: Beyond empirical
  risk minimization. In: ICLR (2018)

\bibitem{ZhouY2014}
Zhou, Y., Chang, H., Barner, K., Spellman, P., Parvin, B.: Classification of
  histology sections via multispectral convolutional sparse coding. In: CVPR
  Workshop. pp. 3081--3088 (2014)

\end{thebibliography}
\end{document}